\documentclass[letterpaper, 10 pt, conference]{ieeeconf}  

\IEEEoverridecommandlockouts                              

\usepackage{graphics} 
\usepackage{epsfig} 
\usepackage{mathptmx} 
\usepackage{times} 
\usepackage{amsmath} 
\usepackage{amssymb}  

\usepackage{bbm}
\usepackage{xurl}
\usepackage[hidelinks]{hyperref}
\usepackage{flafter}
\usepackage{graphicx}
\usepackage{booktabs}
\usepackage{algorithm}
\usepackage{algpseudocode}
\algrenewcommand\algorithmiccomment[1]{\hfill$\triangleright$ #1}
\usepackage{diagbox}
\usepackage{multirow}
\usepackage{tikz}
\usepackage{subcaption}
\newcommand*\circled[1]{\tikz[baseline=(char.base)]{
    \node[shape=circle,draw,inner sep=1pt] (char) {#1};}}
\usepackage{colortbl}
\definecolor{LightCyan}{rgb}{0.88,1,1}
\usepackage{array}
\newcolumntype{C}[1]{>{\centering\arraybackslash}m{#1}}

\title{\LARGE\bfseries
\raisebox{-0.12em}{
    \includegraphics[height=1.3em]{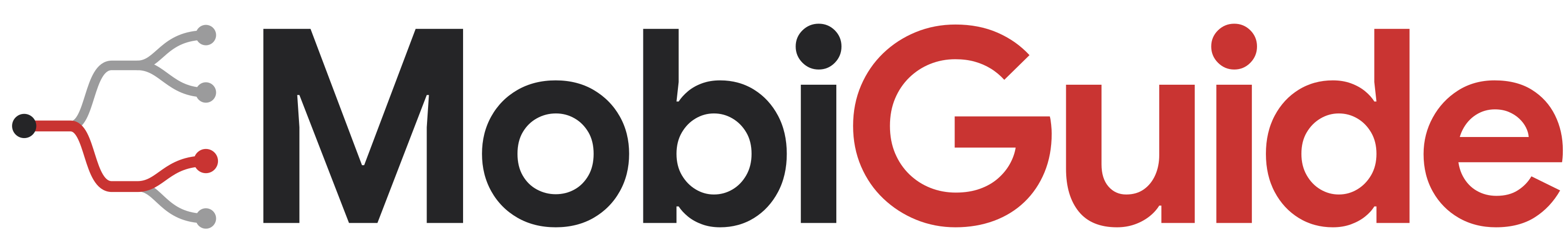}
}
\\[0.05em]
Vocabulary-Guided Online Recovery for Mobile Manipulation
\vspace{0.15em}
}

\author{
Jia Syuen Lim$^{1}$,
Zhizhen Zhang$^{1}$,
Peter Bohm$^{1,2}$,
Brendan Tidd$^{2}$,
Zi Huang$^{1}$,
and Yadan Luo$^{1,*}$%
\thanks{$^{1}$UQMM Lab, The University of Queensland, Brisbane, Australia.
\newline\texttt{\scriptsize
\{jiasyuen.lim, zhizhen.zhang, p.bohm, helen.huang, y.luo\}@uq.edu.au}}%
\thanks{$^{2}$Robotics and Autonomous Systems Group, CSIRO, Brisbane, Australia.
\newline\texttt{\scriptsize brendan.tidd@csiro.au}}%
\thanks{$^{*}$Corresponding author.}%
}
\input{references_compact_setup}  

\begin{document}
\makeatletter

\def\@maketitle{%
  \newpage
  \null
  \vskip 0.5em 
  \begin{center}%
    {\LARGE \bf \@title \par}%
    \vskip 0.7em 
    {\large
      \lineskip .5em%
      \begin{tabular}[t]{c}%
        \@author
      \end{tabular}\par}%
  \end{center}%
  \par
  \vskip 0.3em 
}

\let\@oldmaketitle\@maketitle%
\renewcommand{\@maketitle}{\@oldmaketitle%
\includegraphics[width=\linewidth]{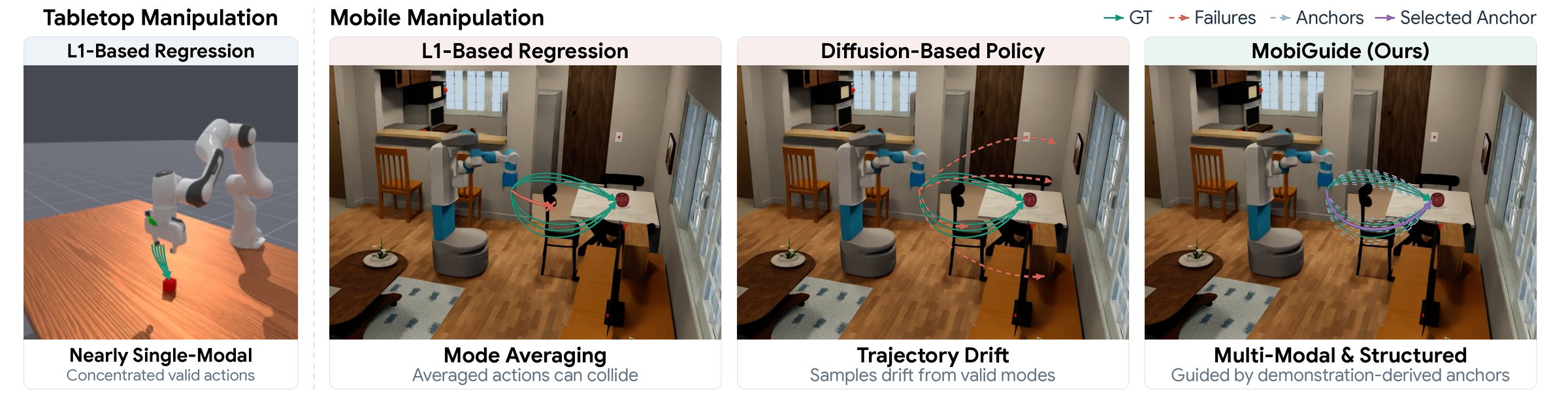}
\refstepcounter{figure}\textbf{Fig.~\thefigure: \textsc{MobiGuide} stabilizes multimodal generation.} Left: Tabletop manipulation is often nearly single-modal, where $L_1$-based VLA regression is sufficient. Right: Mobile manipulation is inherently \textit{multimodal}; $L_1$-based policies average across modes and become infeasible, while diffusion-based policies are more expressive but prone to stochastic drift. \textsc{MobiGuide} guides generation with demonstration-derived anchors, preserving multimodal choices while keeping execution stable.
\label{fig:teaser}
\medskip
\vspace{-10pt}
}

\maketitle
\setcounter{figure}{1}
\thispagestyle{empty}
\pagestyle{empty}

\begin{abstract}

A central challenge in mobile manipulation is preserving multiple plausible action modes while remaining reactive during execution. For example, grasping a bottle in a cluttered scene can be achieved through multiple physically feasible base–arm motion plans. Deterministic VLA policies trained with $L_1$ regression are efficient, but they tend to average across valid modes and may produce physically invalid actions. Diffusion-based generative policies can better represent multimodal action distributions, yet their proposals may be weakly constrained when generated from pure noise, and iterative sampling can increase inference cost. Action chunking helps amortize inference, but it also introduces partially open-loop execution, allowing small mismatches to accumulate into drift. We present \textsc{MobiGuide}, a generative VLA model for efficient mobile manipulation, where action generation is guided by a structured trajectory vocabulary. Constructed from expert demonstrations, this vocabulary encodes plausible whole-body motion patterns and provides structured priors for feasible action generation. Given the current observation and instruction, \textsc{MobiGuide} locally refines these reference trajectories into context-compatible action proposals and selects the most suitable trajectory for execution. This preserves multimodal action candidates while keeping generation constrained around feasible expert-like motion modes. To mitigate chunk-induced drift, we further introduce a lightweight residual correction module that performs high-frequency, per-step adjustments during rollout. Across diverse mobile manipulation tasks, \textsc{MobiGuide} improves success and stability under execution disturbances while maintaining low-latency inference.

\end{abstract}
\section{INTRODUCTION}

Mobile manipulation represents a grand challenge in embodied intelligence, requiring tight integration of semantic reasoning, local base adjustment, and contact-rich arm control for physical interaction near target objects or articulated structures~\cite{DBLP:journals/corr/abs-2505-23692,zhang2025momakitchen100kbenchmarkaffordancegrounded}. The core difficulty lies in the strong temporal coupling between base movement and arm manipulation: early motion decisions shape later viewpoints, object accessibility, and feasible affordances. As a result, even minor state-estimation or actuation errors can propagate over time, progressively shifting the interaction geometry and ultimately leading to task failure. Recent progress in vision-language-action (VLA) models~\cite{Openvla,pi0,pi0.5,rt2,openhelix,openvla-oft,gr00t,smolvla,xvla,lingbotvla,mergevla} has demonstrated the efficacy of mapping visual observations and language instructions directly to actions via large-scale pretraining. However, much of this success has been observed in \textit{tabletop} manipulation settings, where feasible motions are concentrated around a narrow set of solutions (as depicted in Fig.~\ref{fig:teaser}). Even when the robot operates near the target, mobile manipulation is fundamentally less structured.

The same instruction may admit multiple valid whole-body executions, differing in local base pose, approach direction, or grasp geometry, depending on the scene and robot state. This makes the action distribution inherently \textit{multimodal}. Under such supervision, simple teacher-forced ($L_1$) regression often favors an \textbf{average} of several valid behaviors rather than committing to a single executable mode. While these averaged predictions may appear smooth in trajectory space, they are often \textit{invalid} in practice, e.g., by producing collision-prone paths or inaccurate end-effector alignment.

To capture this inherent multimodality, generative policies such as diffusion-based methods have emerged as a powerful alternative to deterministic regression models. These methods can represent multiple plausible action modes rather than collapsing them into a single average prediction. However, state-of-the-art approaches such as AC-DiT~\cite{DBLP:journals/corr/abs-2507-01961}, which directly applies a diffusion transformer (DiT) to mobile manipulation, still face both \textit{generation-level} and \textit{execution-level} challenges. At the generation level, samples from an unstructured prior like pure noise may be diverse but not necessarily concentrated around feasible expert-like whole-body motion patterns, and iterative sampling can increase inference cost. At the execution level, these policies are often deployed through action chunks for efficiency, but the remaining precomputed actions can become \textit{misaligned} as perception evolves, causing small errors to compound into drift when local base motion and arm control are tightly coupled~\cite{mobilealoha,echovla}. While receding-horizon execution and reactive updates can mitigate such chunk-induced drift~\cite{dp,yuan2024unpacking,so2025improving,xue2025reactive,yuan2024policy}, adapting large vision-language backbones to support such behavior remains computationally expensive and often brittle across embodiments and task suites.

These observations motivate a central \textit{question}: can we design a VLA policy for mobile manipulation that is simultaneously \circled{1} computationally efficient, \circled{2} capable of modeling multimodal action distributions, and \circled{3} structured enough to remain executable under chunk-induced drift?

We answer this question with \textbf{\textsc{MobiGuide}}, a generative VLA policy that stabilizes long-horizon mobile manipulation via guided trajectory generation and test-time refinement. The key idea is to leverage training demonstrations to construct a vocabulary of \textit{trajectory anchors} from training demonstrations, where each anchor represents a typical expert motion pattern and provides a structured prior over plausible whole-body behaviors. Given the current observation and instruction, \textsc{MobiGuide} locally refines these reference trajectories into context-compatible action proposals and uses a scoring head to select the most suitable trajectory for execution. This design preserves multimodal action candidates while constraining generation around feasible demonstration-induced motion modes, reducing the burden of unconstrained generative sampling as used in diffusion-based methods. To improve robustness, we introduce a residual correction module for lightweight test-time self-correction, enabling the policy to make per-step adjustments during rollout, compensate for accumulated errors, and recover from deviations without expensive re-planning.

We validate \textsc{MobiGuide} through large-scale simulation and real-world deployment. In simulation, we evaluate on the ManiSkill-HAB mobile manipulation benchmark across six pick, place, open, and close tasks; in the real world, we deploy on a Unitree Go2 Edu quadruped equipped with an SO101 arm, where legged-base vibration and drift provide a challenging test for execution-time correction. On ManiSkill-HAB, \textsc{MobiGuide} achieves the highest average success among the evaluated methods, reaching 66.17\% at $H=2$ and outperforming the strongest point-cloud and RGB baselines, AC-DiT and RDT, by 14.85 and 23.49 percentage points, respectively. With a longer commit length of $H=5$, it maintains 63.39\% average success, exceeding all evaluated baselines. The modest 2.78-point decrease from $H=2$ shows that \textsc{MobiGuide} supports longer action chunks and less frequent replanning while retaining strong task performance.
\section{RELATED WORK}

Recent vision-language-action (VLA) models adapt large-scale vision-language backbones for robotic control, either by representing actions as token sequences~\cite{rt1,rt2,Openvla} or by attaching compact action heads to frozen or partially frozen backbones~\cite{DBLP:journals/corr/abs-2509-09372,smolvla,gr00t,pi0,starvla2025,internvla,evovla}. These approaches improve semantic generalization and parameter efficiency, but long-horizon mobile manipulation additionally requires coordinated prediction of base and arm motions under multimodal task constraints. Existing mobile manipulation systems often address this coordination through structured decomposition, such as base waypoint prediction, viewpoint optimization, or hierarchical planning~\cite{DBLP:conf/cvpr/WuZX0Y25,DBLP:journals/corr/abs-2505-23692}. Other methods learn whole-body trajectory distributions more directly, including scene-conditioned trajectory optimization in M2Diffuser~\cite{DBLP:journals/corr/abs-2410-11402} and diffusion-style whole-body action modeling in AC-DiT~\cite{DBLP:journals/corr/abs-2507-01961}.

In parallel, diffusion policies have been used to capture multimodal action distributions in robotic manipulation and planning~\cite{rdt,scaledp}, while some trajectory priors have been explored for efficient refinement in autonomous driving and manipulation~\cite{DBLP:conf/cvpr/LiaoCY0WYZLZZW25, DBLP:conf/cvpr/XingZHJHZLY25, DBLP:journals/corr/abs-2506-06664, DBLP:journals/corr/abs-2503-12030, DBLP:journals/corr/abs-2406-06978,anchordp3}.

Our method follows this broader principle, combining a vocabulary of demonstration-derived whole-body action chunks with context-conditioned candidate refinement and scoring, followed by per-step residual correction during execution. Our focus is the success--latency trade-off of this combination for chunked mobile manipulation.

\section{METHODOLOGY}
\label{sec:method}

\subsection{Problem Formulation}
\label{subsec:problem_formulation}

We consider language-conditioned mobile manipulation, where a robot with a mobile base and a manipulator must execute long-horizon tasks in unstructured environments. At timestep $t$, the robot observes $o_t=\{I_t^g,I_t^w,s_t\}$, where $I_t^g$ and $I_t^w$ are global- and wrist-view RGB observations, and $s_t$ denotes proprioceptive states including the base state, arm joint states, end-effector pose, and gripper status. Given a language instruction $\ell$, the policy predicts a future action chunk
\begin{equation}
A_t = a_{t:t+T-1} \in \mathbb{R}^{T\times d},
\end{equation}
where $T$ is the prediction horizon and each action $a_t\in\mathbb{R}^d$ jointly controls the mobile base and manipulator. During execution, the policy commits to the first $H\leq T$ actions before replanning, where $H$ denotes the commit length.
\begin{figure}[t]
  \centering
  \includegraphics[width=\columnwidth]{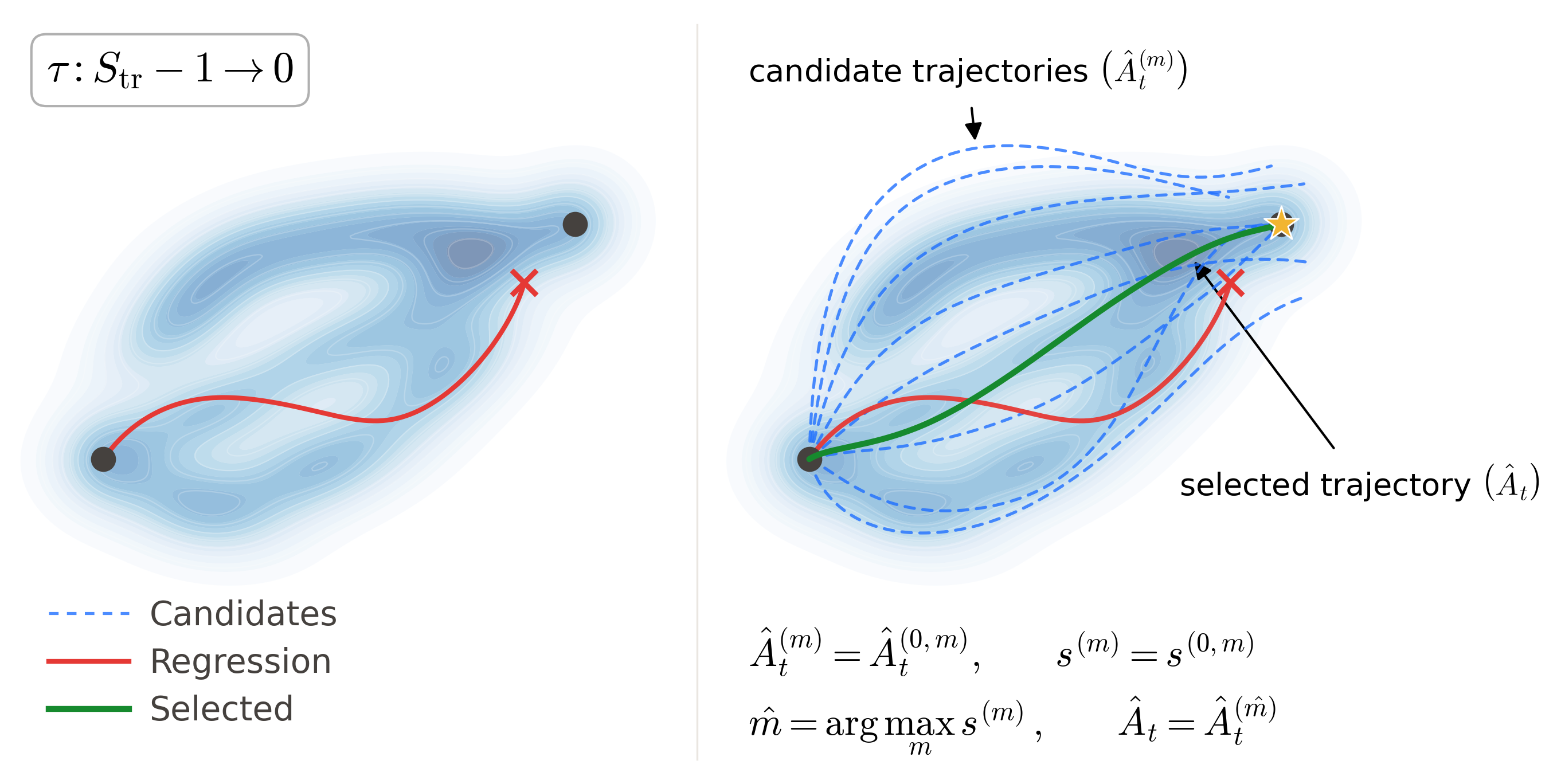}
  \caption{Comparison between uni-modal prediction and multimodal vocabulary-guided local refinement. \textsc{MobiGuide} starts from a demonstration-derived vocabulary and refines locally, constraining generation to plausible actions.}
  \label{fig:l1_vs_mobiguide}
\end{figure}
\begin{figure}[!t]
    \centering
    \includegraphics[width=\columnwidth]{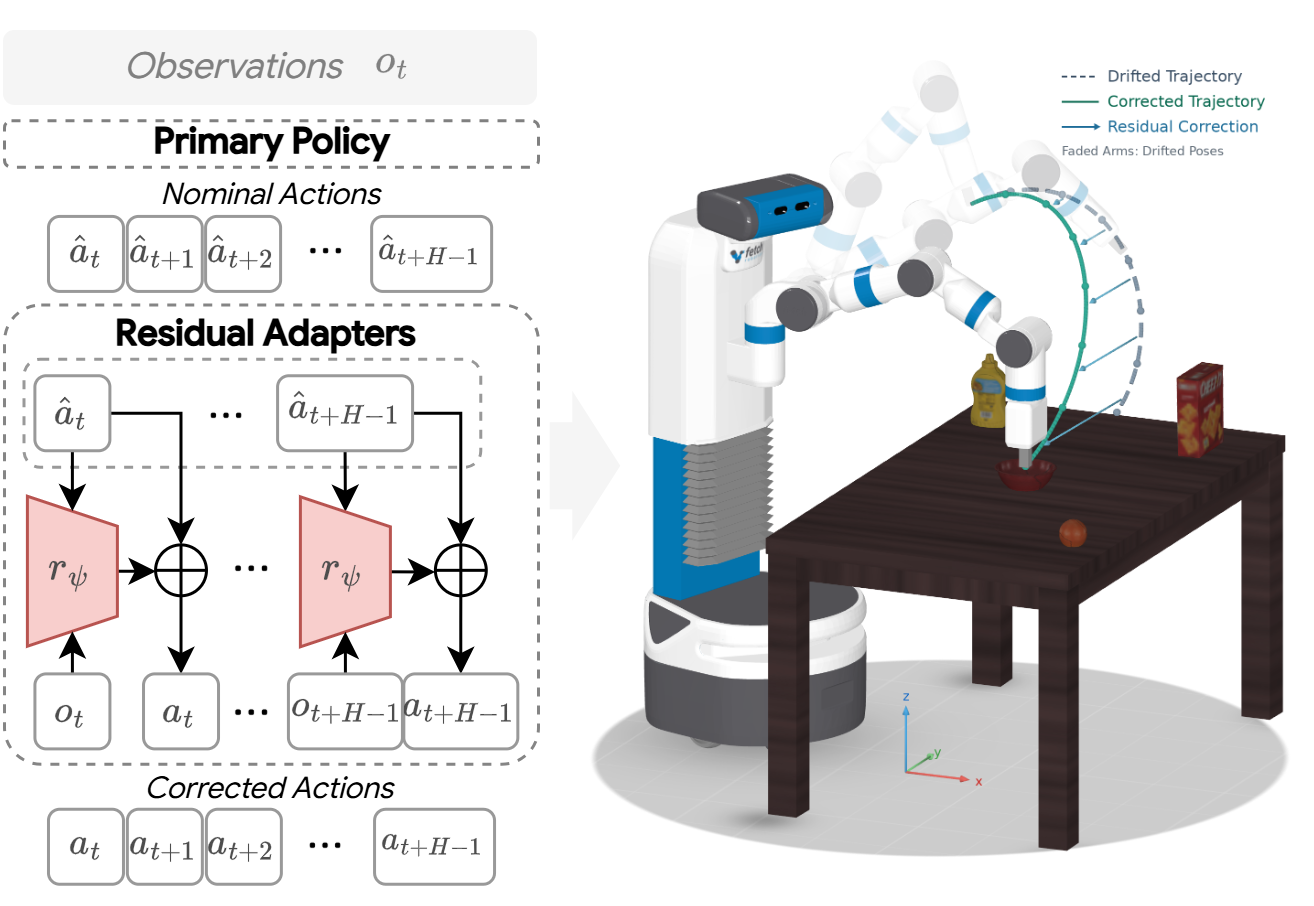}
    \caption{During chunked execution, residual correction predicts state-dependent micro-adjustments to compensate for base drift and arm--object misalignment without repeatedly querying the full VLA policy.}
    \label{fig:residual}
\end{figure}
\subsection{Framework Overview}
\label{subsec:framework_overview}
\begin{figure*}[t]
    \centering
    \includegraphics[width=\textwidth]{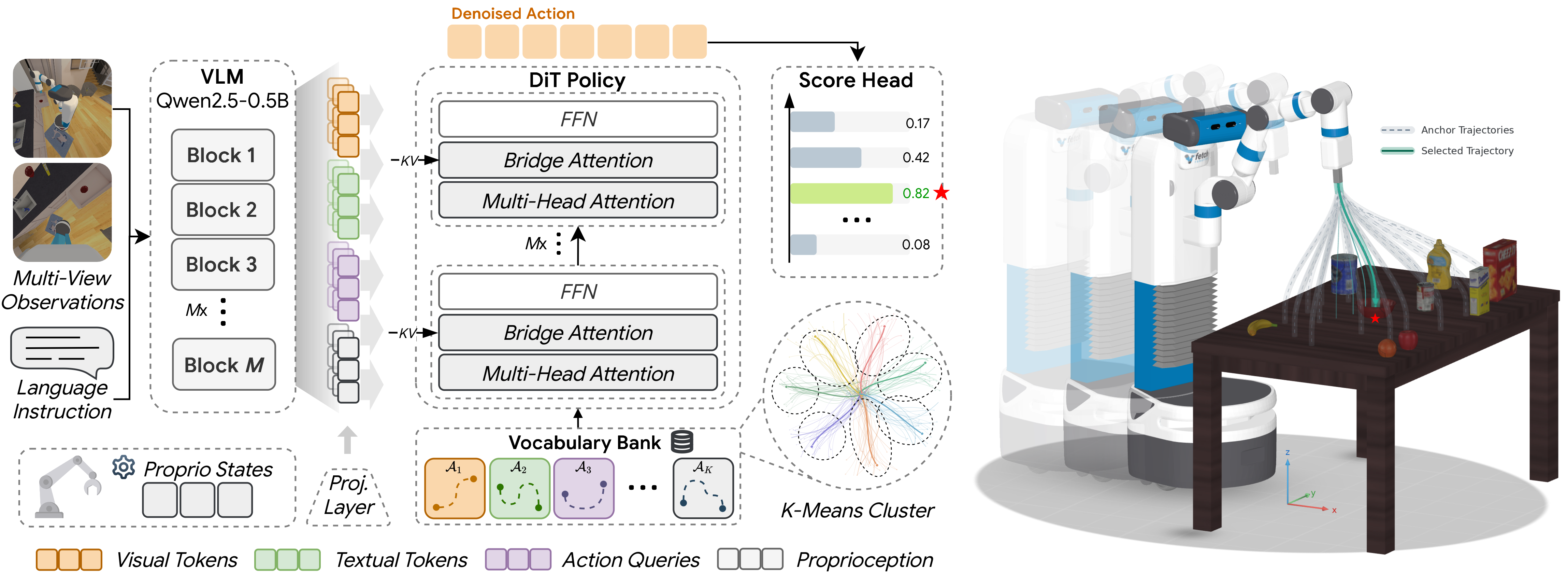}
    \caption{\textbf{Overview of the \textsc{MobiGuide} framework.} The VLA backbone encodes multimodal context. The action head refines demonstration-derived trajectory anchors into context-conditioned proposals, and a scoring head selects the highest-scoring context-compatible proposal.}
    \label{fig:MobiGuide}
\end{figure*}
Fig.~\ref{fig:MobiGuide} presents an overview of \textsc{MobiGuide}. The framework addresses the trade-off between efficient deterministic regression and multimodal generative modeling. While $L_1$ action heads are simple and low-latency, they can suffer from \textit{mode averaging} when multiple valid action chunks exist. Generative policies better capture such multimodality, but sampling from broad noise priors can produce \textit{weakly constrained proposals} and increase inference cost. \textsc{MobiGuide} resolves this through \textbf{vocabulary-guided local refinement}: instead of regressing a single action chunk or generating from pure noise, it refines demonstration-derived trajectory anchors into context-compatible proposals and scores them for execution. This preserves multimodal candidates while restricting refinement to neighborhoods around plausible expert trajectories.

Concretely, \textsc{MobiGuide} contains three components. First, a lightweight context encoder combines VLM features from the visual-language inputs with proprioceptive state embeddings:
\begin{equation}
    x_t =
    \Phi_{\mathrm{ctx}}
    \left(
        f_{\mathrm{VLM}}(I_t^g, I_t^w, \ell),
        f_{\mathrm{state}}(s_t)
    \right),
\end{equation}
where $x_t$ summarizes the visual observations, proprioceptive state, and language-conditioned task information. Second, the anchor-guided action head conditions on $x_t$ to refine and score multiple trajectory anchors. Third, during rollout, a lightweight residual correction module applies step-wise action adjustments to reduce execution drift during chunked control.
\subsection{Trajectory Anchors as Structured Action Priors}
\label{subsec:trajectory_anchors}

\textsc{MobiGuide} represents expert motion patterns using a vocabulary of trajectory anchors:
\begin{equation}
    \mathcal{V}=\{\bar{A}^{(m)}\}_{m=1}^{M},
    \qquad
    \bar{A}^{(m)} \in \mathbb{R}^{T \times d},
\end{equation}
where $T$ is the predicted chunk length and $d$ is the action dimension.

We construct $\mathcal{V}$ offline by segmenting expert demonstrations into fixed-length action chunks and using their K-means cluster centers as anchors. Before clustering, the 13-dimensional whole-body actions are normalized to $[-1,1]$ to reduce the effects of scale differences across dimensions. We use $M=20$ in our primary experiments; Fig.~\ref{fig:k_sweep} evaluates the success--latency trade-off across vocabulary sizes.

These anchors provide trajectory priors for context-conditioned refinement. Conceptually, the conditional action distribution is approximated by an anchor-conditioned mixture:
\begin{equation}
    p(A_t \mid x_t)
    \approx
    \sum_{m=1}^{M}
    p_\theta(m \mid x_t)
    p_\theta(A_t \mid x_t,\bar{A}^{(m)}).
\end{equation}
Here, $p_\theta(m \mid x_t)$ represents relative anchor--context compatibility, while $p_\theta(A_t \mid x_t,\bar{A}^{(m)})$ models local adaptation around each anchor. At inference, the policy refines and scores the candidates and executes the highest-scoring proposal.
\subsection{Anchor-Guided Local Refinement}
\label{subsec:anchor_guided_refinement}

Given the context representation $x_t$ and anchor vocabulary $\mathcal{V}$, the action head refines each anchor locally rather than generating an action chunk from pure noise. Let $S$ denote the full number of refinement levels and $\rho\in(0,1]$ the truncation ratio. We define the truncated refinement horizon and its low-noise levels as
\begin{equation}
    S_{\mathrm{tr}} = \max\{1,\lfloor \rho S \rceil\},
    \qquad
    \tau \in \{0,\dots,S_{\mathrm{tr}}-1\}.
\end{equation}

For each anchor $\bar{A}^{(m)}$, we construct an anchor-local perturbed trajectory at refinement level $\tau$:
\begin{equation}
    A_t^{(\tau,m)}
    =
    \sqrt{\bar{\alpha}_{\tau}}\bar{A}^{(m)}
    +
    \sqrt{1-\bar{\alpha}_{\tau}}\epsilon,
    \qquad
    \epsilon \sim \mathcal{N}(0,I).
    \label{eq:anchor_perturb}
\end{equation}
The schedule controls the perturbation radius: larger $\tau$ explores a broader anchor-local neighborhood, while smaller $\tau$ enables fine-grained correction.

The perturbed trajectory is processed by a DiT-style action transformer conditioned on the context, anchor, and timestep embedding $e_\tau=\phi_{\mathrm{time}}(\tau)$:
\begin{equation}
    z_t^{(\tau,m)}
    =
    T_\theta\!\left(
        A_t^{(\tau,m)},
        x_t,
        e_\tau,
        \bar{A}^{(m)}
    \right).
\end{equation}
A refinement head predicts a residual and produces the updated action proposal:
\begin{equation}
    \begin{aligned}
        \Delta A_t^{(\tau,m)}
        &=
        r_\theta\!\left(z_t^{(\tau,m)}\right), \\
        \hat{A}_t^{(\tau,m)}
        &=
        A_t^{(\tau,m)}+\Delta A_t^{(\tau,m)}.
    \end{aligned}
\end{equation}
A separate scoring head estimates the proposal's compatibility with the current context:
\begin{equation}
    s^{(\tau,m)}
    =
    h_\theta\!\left(z_t^{(\tau,m)}\right).
\end{equation}

At inference, all anchors are evaluated in parallel. For each anchor, we initialize the trajectory at the largest truncated refinement level, $\tau_{\mathrm{start}}=S_{\mathrm{tr}}-1$, using Eq.~\eqref{eq:anchor_perturb}. We then iteratively refine the trajectory toward $\tau=0$, applying
\begin{equation}
    A_t^{(\tau-1,m)}
    =
    \mathcal{U}_{\tau}\!\left(
        A_t^{(\tau,m)},
        \hat{A}_t^{(\tau,m)}
    \right)
\end{equation}
for $\tau=\tau_{\mathrm{start}},\dots,1$, where $\mathcal{U}_{\tau}$ denotes the truncated reverse update from level $\tau$ to $\tau-1$.

After the final refinement step, each anchor yields a locally refined proposal and a compatibility score. The policy selects the highest-scoring proposal as the final macro action chunk (Fig.~\ref{fig:l1_vs_mobiguide}), adapting demonstration-derived motion patterns to the current context through truncated local refinement.
\subsection{Test-time Residual Correction for Chunked Execution}
\label{subsec:residual_correction}
After selecting the nominal action chunk $\hat{A}_t=\hat{a}_{t:t+T-1}$, the robot executes its first $H\leq T$ actions before replanning. During execution, small base deviations can alter the manipulator--object geometry. To mitigate this drift without querying the full VLA policy at every step, we apply
\begin{equation}
    a_{t+j}
    =
    \hat{a}_{t+j}
    +
    r_\psi(o_{t+j},\ell,\hat{a}_{t+j},j),
    \label{eq:residual_action}
\end{equation}
for $j\in\{0,\dots,H-1\}$, where $r_\psi$ predicts a state-dependent residual using the updated observation.

\noindent\textbf{Implementation and Training.} We implement $r_\psi$ as a lightweight MLP without an explicit history of observations or actions. A 1D sinusoidal embedding of $j$ is concatenated with the visual and action features. After training the primary \textsc{MobiGuide} policy, we freeze its parameters and predict an action chunk from each recorded expert-demonstration observation $o_t$, taking the action at offset $j$ as the nominal action $\hat{a}_{t+j}$. During residual training, $o_{t+j}$ and $a^\star_{t+j}$ are the recorded observation and paired expert action at timestep $t+j$ from the same demonstration. We train $r_\psi$ on these demonstration observations with target residuals $\Delta a^\star_{t+j}=a^\star_{t+j}-\hat{a}_{t+j}$ by minimizing
\begin{equation}
    \mathcal{L}_{\mathrm{res}}
    =
    \mathbb{E}\!\left[
        \left\|
        r_\psi(o_{t+j},\ell,\hat{a}_{t+j},j)
        -
        \Delta a^\star_{t+j}
        \right\|_1
    \right].
    \label{eq:residual_loss}
\end{equation}
\subsection{Learning Objective}
\label{subsec:learning_objective}

During training, each expert action chunk $A_t^\star$ is assigned to the closest anchor under a trajectory-level distance metric:
$
    m^\star
    =
    \arg\min_m
    d
    \left(
        A_t^\star,
        \bar{A}^{(m)}
    \right),
$
where $d(\cdot,\cdot)$ measures the distance between two action chunks. The corresponding anchor label is defined as $y_m=\mathbbm{1}[m=m^\star]$. We jointly train the refinement and scoring heads with:
\begin{equation}
    \mathcal{L}
    =
    \sum_{m=1}^{M}
    \left[
        y_m
        \,
        \mathcal{L}_{\mathrm{ref}}
        \left(
            \hat{A}_t^{(m)},
            A_t^\star
        \right)
        +
        \lambda
        \,
        \mathcal{L}_{\mathrm{score}}
        \left(
            s^{(m)}, y_m
        \right)
    \right].
    \label{eq:anchor_loss}
\end{equation}
where $\mathcal{L}_{\mathrm{ref}}$ is a trajectory reconstruction loss over the action chunk, and $\mathcal{L}_{\mathrm{score}}$ is a binary classification loss for anchor compatibility.
\vspace{-1ex}
\begin{figure*}[t]
    \centering
    \includegraphics[width=\textwidth]{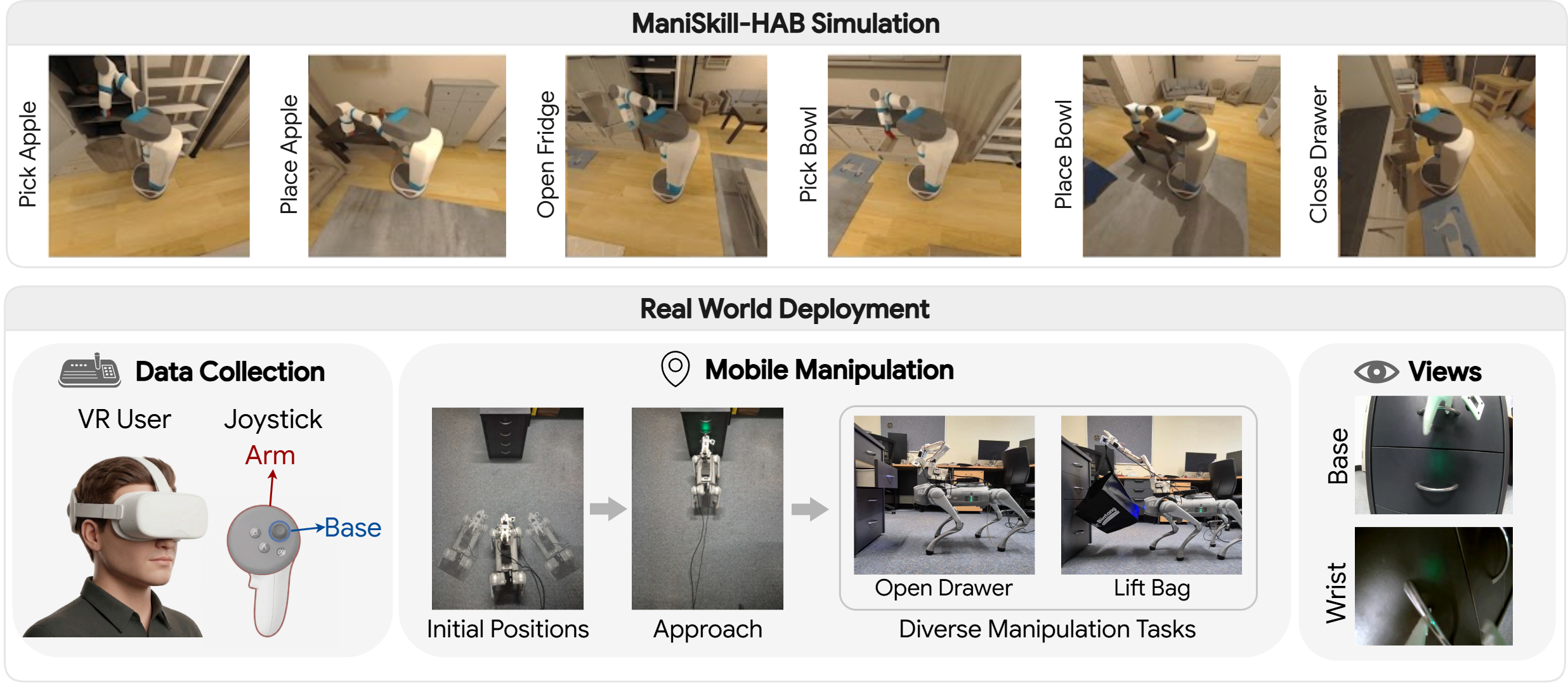}
    \caption{Experimental Environment. Simulation experiments are conducted in ManiSkill-HAB. Real-world experiments are evaluated on the Unitree Go2 quadruped equipped with an SO101 arm. The tasks need the robot to first navigate to the target place, then interact with the objects. The perception module includes two camera views, \textit{i.e.,} base view from Go2 and wrist view from SO101. The data collection process was conducted by teleoperation with the Meta Quest 3 VR device.}
    \label{fig:real_robot}
\end{figure*}

\section{Experiments}
\label{sec:exp}
\subsection{Experimental Setup}
\textbf{Environments and Tasks.} We conduct empirical evaluations within the ManiSkill-HAB (MS-HAB) framework \cite{DBLP:conf/iclr/ShuklaTS25}, built on the SAPIEN physics engine~\cite{DBLP:conf/cvpr/XiangQMXZLLJYWY20}. Our simulation suite uses the \textbf{SetTable} scenario with six mobile manipulation tasks: \textsc{Pick Apple}, \textsc{Place Apple}, \textsc{Pick Bowl}, \textsc{Place Bowl}, \textsc{Open Fridge}, and \textsc{Close Drawer Counter}. To assess real-world applicability, we deploy \textsc{MobiGuide} on a Unitree Go2 Edu quadruped equipped with an SO101 arm, using two representative tasks: \textsc{Open Drawer} and \textsc{Lift Bag}. As shown in Fig.~\ref{fig:real_robot}, each task requires navigation to the target followed by manipulation. We collect 50 teleoperated demonstrations per task.

\noindent \textbf{Dataset and Training Setup.} \textsc{MobiGuide} uses Qwen2.5-0.5B~\cite{qwen2025qwen25technicalreport} as VLM backbone for efficiency. Our training regime leverages expert trajectories generated by reinforcement learning algorithms \cite{DBLP:conf/iclr/ShuklaTS25, DBLP:journals/corr/SchulmanWDRK17, DBLP:journals/corr/abs-1812-05905}, yielding a fixed dataset of 1,000 successful rollouts per task. The robot's observation is formulated as a single-timestep multimodal input. At each step $t$, the policy processes RGB images from the head and wrist cameras alongside a dense proprioceptive and task-state vector. This vector explicitly captures the base state (pose, linear, and angular velocities), the manipulator's kinematics (joint positions and velocities), the end-effector pose, the gripper status, and the target object or goal pose. Based on this single-frame context, the model outputs a unified 13-DoF continuous action command. To adapt the model efficiently, we apply LoRA~\cite{hu2021loralowrankadaptationlarge} to the VLM backbone while fully fine-tuning the lightweight DiT action head.

\noindent\textbf{Baselines.} Ensuring technical reproducibility in mobile manipulation is frequently hindered by the limited availability of open-source codebases for recent systems. Consequently, we benchmark \textsc{MobiGuide} against six publicly accessible, state-of-the-art architectures, logically grouped into three core paradigms. First, we compare against foundational imitation learning frameworks: Action Chunking with Transformers (ACT) \cite{DBLP:conf/rss/ZhaoKLF23} and Diffusion Policy (DP) \cite{dp}. Second, we evaluate against geometrically-aware and large-scale robotic foundation models, including 3D Diffusion Policy (DP3) \cite{DBLP:conf/rss/ZeZZHWX24} for explicit 3D visual processing, alongside highly scalable generalist models like $\pi_0$ \cite{pi0} and Robotics Diffusion Transformer (RDT) \cite{rdt}. Finally, we benchmark against AC-DiT \cite{DBLP:journals/corr/abs-2507-01961}, a recently proposed, state-of-the-art diffusion transformer explicitly tailored for the coupled dynamics of mobile manipulation.

\noindent\textbf{Input Protocol.} All methods receive the same proprioceptive and task-state information, including object and goal poses. The visual modality differs: RGB-based methods, including \textsc{MobiGuide}, receive RGB observations, whereas DP3 and AC-DiT receive point clouds.

\noindent \textbf{Evaluation Metrics.} We quantify policy performance using the task success rate. An evaluation episode in simulation is strictly defined as successful if the agent satisfies all designated goal conditions within a maximum execution horizon of 200 timesteps. For rigorous statistical validation, we evaluate each method across three independent trials. During each trial, we execute 100 unseen episodes per task. We report the final performance as the mean and standard deviation of the success rates across these three independent trials.

\begin{table*}[t]
\centering
\caption{ \textbf{Simulation results and residual correction ablation on ManiSkill-HAB.} We report Success Once (\%, mean $\pm$ standard deviation across repeated evaluations). Visual Input denotes the visual observation modality; all methods additionally receive the shared proprioceptive and task-state inputs described in Sec.~\ref{sec:exp}-A. $H$ denotes the commit length. For \textsc{MobiGuide}, $\dagger$ denotes variants without residual correction; unmarked variants include residual correction. Average is the unweighted mean across the six tasks. Bold indicates the highest mean in each column.
}
\label{tab:mshab_main}

\resizebox{\textwidth}{!}{
\begin{tabular}{ccc cccccc c}
\toprule
\multirow{3}{*}{\textbf{Methodology}}
& \multirow{3}{*}{\textbf{Venue}}
& \multirow{3}{*}{\textbf{Visual Input}}
& \multicolumn{2}{c}{\textsc{Pick}}
& \multicolumn{2}{c}{\textsc{Place}}
& \textsc{Open}
& \textsc{Close}
& \multirow{3}{*}{\textbf{Average}} \\
\cmidrule(lr){4-5}
\cmidrule(lr){6-7}
\cmidrule(lr){8-8}
\cmidrule(lr){9-9}
& &
& Apple & Bowl & Apple & Bowl & Door & Drawer & \\
& &
& Fridge & Counter & Table & Table & Fridge & Counter & \\
\midrule

$\pi_{0}$
& -- & RGB
& $13.0 \pm 1.6$ & $15.7 \pm 1.2$
& $23.3 \pm 1.7$ & $21.3 \pm 2.6$
& $31.3 \pm 2.6$ & $70.0 \pm 2.2$
& 29.10 \\

\textsc{ACT}
& RSS'23 & RGB
& $28.0 \pm 2.2$ & $28.0 \pm 2.4$
& $8.7 \pm 3.3$ & $13.0 \pm 0.8$
& $2.0 \pm 2.2$ & $85.7 \pm 1.2$
& 27.57 \\

\textsc{RDT}
& ICLR'25 & RGB
& $12.0 \pm 11.3$ & $10.7 \pm 6.8$
& $32.0 \pm 5.7$ & $18.7 \pm 5.0$
& $82.7 \pm 10.5$ & $\mathbf{100.0 \pm 0.0}$
& 42.68 \\

\textsc{DP}
& RSS'23 & RGB
& $21.3 \pm 3.3$ & $20.7 \pm 3.3$
& $28.0 \pm 8.0$ & $69.3 \pm 3.3$
& $7.3 \pm 5.8$ & $55.0 \pm 5.7$
& 33.60 \\

\textsc{DP3}
& RSS'24 & PointCloud
& $0.0 \pm 0.0$ & $20.0 \pm 2.4$
& $31.0 \pm 0.8$ & $32.0 \pm 0.8$
& $0.0 \pm 0.0$ & $68.0 \pm 0.0$
& 25.17 \\

\textsc{AC-DiT}
& NeurIPS'25 & PointCloud
& $33.3 \pm 1.9$ & $36.0 \pm 6.5$
& $33.3 \pm 9.4$ & $17.3 \pm 6.8$
& $\mathbf{90.7 \pm 5.0}$ & $97.3 \pm 1.9$
& 51.32 \\

\midrule

\rowcolor{LightCyan!40}
\textsc{MobiGuide}$_{H=2}^{\dagger}$
& Ours & RGB
& $20.33 \pm 0.58$ & $39.00 \pm 2.00$
& $61.33 \pm 0.58$ & $70.67 \pm 0.58$
& $77.00 \pm 1.00$ & $92.00 \pm 1.00$
& 60.06 \\

\rowcolor{LightCyan!40}
\textsc{MobiGuide}$_{H=2}$
& Ours & RGB
& $\mathbf{34.00 \pm 2.65}$ & $\mathbf{50.67 \pm 1.53}$
& $62.33 \pm 0.58$ & $\mathbf{79.67 \pm 0.58}$
& $74.00 \pm 0.00$ & $96.33 \pm 0.58$
& \textbf{66.17} \\

\arrayrulecolor{gray}
\midrule
\arrayrulecolor{black}

\rowcolor{LightCyan!40}
\textsc{MobiGuide}$_{H=5}^{\dagger}$
& Ours & RGB
& $15.67 \pm 0.58$ & $32.67 \pm 0.58$
& $\mathbf{63.67 \pm 0.58}$ & $73.00 \pm 0.00$
& $53.67 \pm 0.58$ & $92.67 \pm 0.58$
& 55.23 \\

\rowcolor{LightCyan!40}
\textsc{MobiGuide}$_{H=5}$
& Ours & RGB
& $22.33 \pm 1.15$ & $40.33 \pm 2.31$
& $63.00 \pm 1.00$ & $79.00 \pm 0.00$
& $77.00 \pm 0.00$ & $98.67 \pm 0.58$
& 63.39 \\

\bottomrule
\end{tabular}
}
\end{table*}

\begin{table}[t]
\centering
\caption{Success rate comparison between \textsc{MobiGuide} and baselines on real-world tasks.}
\label{tab:real_table}
\resizebox{\columnwidth}{!}{
\begin{tabular}{lccc}
\toprule
\textbf{Method} & \textbf{Open Drawer} & \textbf{Lift Bag} & \textbf{Average} \\
\midrule
\textsc{ACT} & 10\% & 0\% & 5\% \\
\textsc{Diffusion Policy} & 20\% & 0\% & 10\% \\
\textsc{SmolVLA} & 20\% & 10\% & 15\% \\
\rowcolor{LightCyan!40}
\textsc{MobiGuide} & \textbf{70\%} & \textbf{50\%} & \textbf{60\%} \\
\bottomrule
\end{tabular}
}
\end{table}
\subsection{Results on ManiSkill-HAB}
Table~\ref{tab:mshab_main} reports task success rates on the ManiSkill-HAB mobile manipulation benchmark across \textsc{Pick}, \textsc{Place}, \textsc{Open}, and \textsc{Close} tasks. We compare \textsc{MobiGuide} against strong RGB-based baselines ($\pi_0$, ACT, RDT, DP) and point-cloud-based baselines (DP3, AC-DiT), with all baselines evaluated using a short commit length of $H=2$ for responsive control. Without using point-cloud observations, \textsc{MobiGuide}$_{H=2}$ achieves an average success rate of 66.17\%, outperforming the strongest point-cloud baseline, AC-DiT, by 14.85 percentage points. Notably, \textsc{MobiGuide}$_{H=2}$ obtains large gains on placement tasks, improving over RDT on \textsc{Place Apple} by 30.33 points and over AC-DiT on \textsc{Place Bowl} by 62.37 points, while achieving 96.33\% success on \textsc{Close Drawer}.

We further evaluate \textsc{MobiGuide} with a longer commit length of $H=5$ to test whether its action chunks remain reliable under less frequent replanning. The $H=5$ variant achieves an average success rate of 63.39\%, exceeding all baselines evaluated at $H=2$ and outperforming AC-DiT by 12.07 percentage points. Compared with \textsc{MobiGuide}$_{H=2}$, the average success rate decreases by 2.78 points. These results suggest that anchor-conditioned dense trajectory reconstruction supports reliable multi-step action chunks under less frequent replanning, reducing the frequency of policy inference while retaining strong overall task performance.

\subsection{Results on Real World Robot}
To evaluate the effectiveness of our method, we compare against three baseline models, including ACT~\cite{DBLP:conf/rss/ZhaoKLF23}, DP~\cite{dp} and one VLA model with a comparable parameter scale, SmolVLA-0.5B~\cite{smolvla}. Performance is measured by the task success rate reported in Table~\ref{tab:real_table}. Each task is evaluated over 10 trials with randomized initial positions. Our method achieves consistently higher success rates across both tasks. In particular, \textsc{MobiGuide} improves the overall success rate from 15\% to 60\%, demonstrating a significant improvement in mobile manipulation behaviors under real-world conditions.

\subsection{Ablation Study}
To isolate the specific contributions of our proposed architectural choices, we conduct targeted ablation studies within the ManiSkill-HAB benchmark suite, specifically focusing our evaluation on the complex \textsc{Pick Bowl} task.

\begin{table}[t]
    \centering
    \caption{Anchor-based and Gaussian-initialized generation under different denoising-step settings on \textsc{Pick Bowl}. Values are mean $\pm$ standard deviation across three trials. Latency measures a complete batch-one action-chunk prediction, including preprocessing, VLM inference, and diffusion. All experiments use $H=5$ and residual correction is disabled.}
    \label{tab:anchor_prior_ablation}
    \resizebox{\columnwidth}{!}{
    \begin{tabular}{lccc}
        \toprule
        \textbf{Method} &
        \textbf{Steps} &
        \textbf{SR (\%) $\uparrow$} &
        \textbf{Latency (ms) $\downarrow$} \\
        \midrule
        w/o Anchor & 50 & $4.67 \pm 0.58$ & $938.37 \pm 2.48$ \\
        w/o Anchor & 25 & $4.00 \pm 0.00$ & $501.09 \pm 2.34$ \\
        w/o Anchor & 10 & $5.67 \pm 0.58$ & $244.91 \pm 1.76$ \\
        \midrule
        \rowcolor{LightCyan!40}
        w/ Anchor & 10 & $\mathbf{32.67 \pm 0.58}$ & $249.72 \pm 2.89$ \\
        \rowcolor{LightCyan!40}
        w/ Anchor & 5 & $26.00 \pm 1.73$ & $\mathbf{155.81 \pm 2.26}$ \\
        \bottomrule
    \end{tabular}}
\end{table}

\begin{table}[t]
      \centering
      \caption{Success Once (\%) under action attenuation at $H=5$, without ($-$) and with ($+$) residual correction. Bold indicates higher success within each pair.}
      \label{tab:attenuation}
      \small
      \setlength{\tabcolsep}{2pt}
      \renewcommand{\arraystretch}{1.12}

      \begin{tabular*}{0.92\linewidth}{
          @{\extracolsep{\fill}}lcccccc@{}
      }
          \toprule
          \multirow{2}{*}{\textbf{Task}}
          & \multicolumn{2}{c}{\textbf{0\%}}
          & \multicolumn{2}{c}{\textbf{10\%}}
          & \multicolumn{2}{c}{\textbf{20\%}} \\
          \cmidrule(lr){2-3}
          \cmidrule(lr){4-5}
          \cmidrule(lr){6-7}
          & $-$ & $+$ & $-$ & $+$ & $-$ & $+$ \\
          \midrule

          Pick Apple
          & 15.7 & \textbf{22.3}
          & 11.7 & \textbf{17.0}
          & 11.0 & \textbf{17.7} \\

          Pick Bowl
          & 32.7 & \textbf{40.3}
          & 20.7 & \textbf{33.0}
          & 11.3 & \textbf{16.0} \\

          \addlinespace[2pt]
          \midrule

          Place Apple
          & \textbf{63.7} & 63.0
          & 64.0 & \textbf{68.7}
          & 61.0 & \textbf{63.7} \\

          Place Bowl
          & 73.0 & \textbf{79.0}
          & 73.3 & \textbf{75.3}
          & 65.3 & \textbf{67.7} \\

          \addlinespace[2pt]
          \midrule

          Open Fridge
          & 53.7 & \textbf{77.0}
          & 38.7 & \textbf{72.7}
          & 29.7 & \textbf{53.0} \\

          Close Counter
          & 92.7 & \textbf{98.7}
          & 91.0 & \textbf{95.7}
          & 82.7 & \textbf{95.0} \\

          \bottomrule
      \end{tabular*}
      \vspace{-3ex}
  \end{table}

\noindent \textbf{Impact of Anchored Generation.} We compare anchor-based generation against a Diffusion Transformer baseline initialized from Gaussian noise $\mathcal{N}(0,I)$. At 10 denoising steps, the Gaussian baseline obtains $5.67 \pm 0.58$\% success, while \textsc{MobiGuide} achieves $32.67 \pm 0.58$\%. The latter also exceeds the Gaussian baseline's success of $4.67 \pm 0.58$\% at 50 steps. Reducing anchored refinement to five steps yields $26.00 \pm 1.73$\% success, outperforming all evaluated Gaussian settings while achieving the lowest measured batch-one latency of $155.81 \pm 2.26$\,ms. These results support the effectiveness of anchor-based generation under reduced denoising budgets. However, because \textsc{MobiGuide} refines and scores multiple candidates, denoising-step counts alone do not establish equivalent computational cost.

\begin{figure*}[h]
    \centering
    \includegraphics[width=\textwidth]{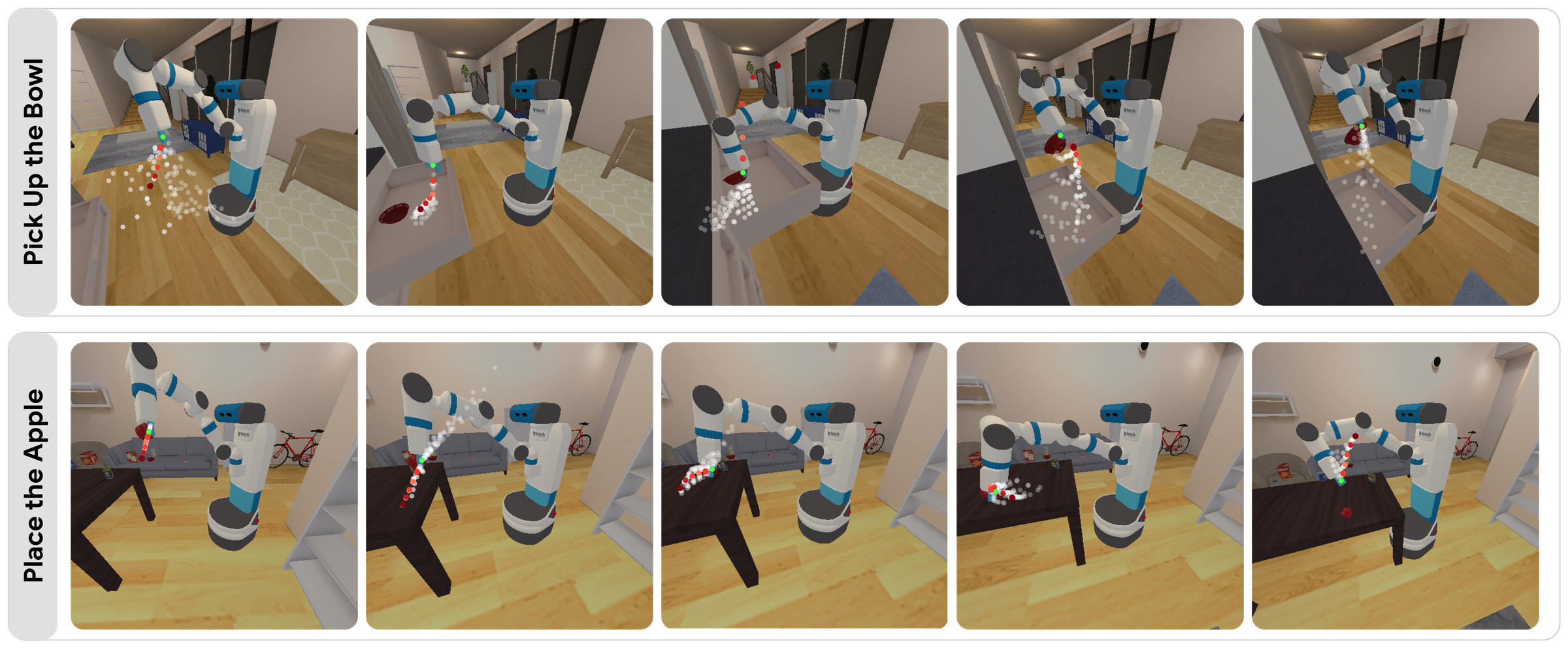}
    \caption{\textbf{Qualitative visualization of \textsc{MobiGuide} in ManiSkill-HAB.} We show task progress from left to right for \textsc{Pick Bowl} and \textsc{Place Apple}. White points denote candidate future end-effector trajectories reconstructed from anchored proposals, while colored points indicate the trajectory selected by the scoring head for execution.}
    \label{fig:demo_vis}
    \vspace{-3ex}
\end{figure*}

\begin{figure}[t]
  \centering
  \includegraphics[width=\columnwidth]{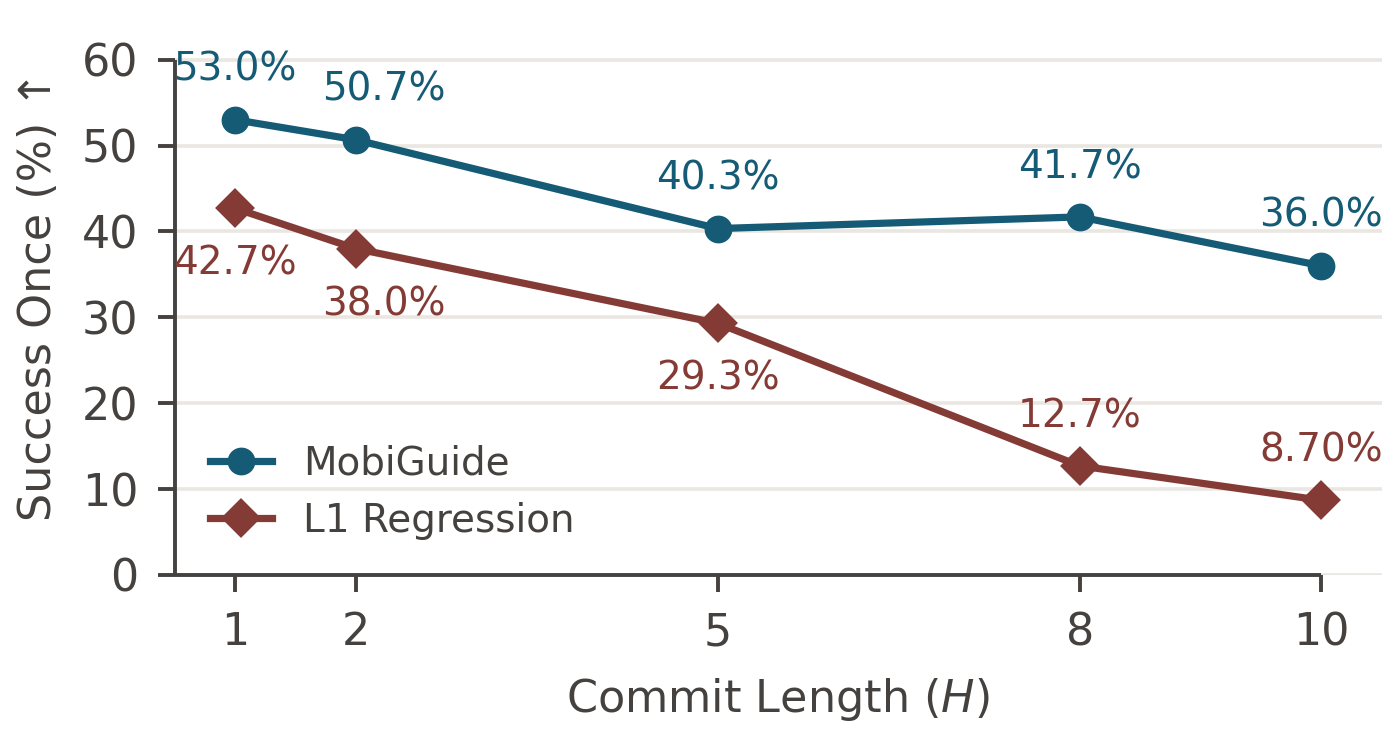}
  \caption{Effect of commit length $H$ on Pick Bowl success. \textsc{MobiGuide} outperforms $L_1$ regression across all settings, with a smaller overall decline from $H=1$ to $H=10$ (53.0\% to 36.0\%, versus 42.7\% to 8.7\%).}
  \label{fig:chunk_impact}
  \vspace{-3ex}
\end{figure}

\begin{figure}[t]
  \centering
  \includegraphics[width=\columnwidth]{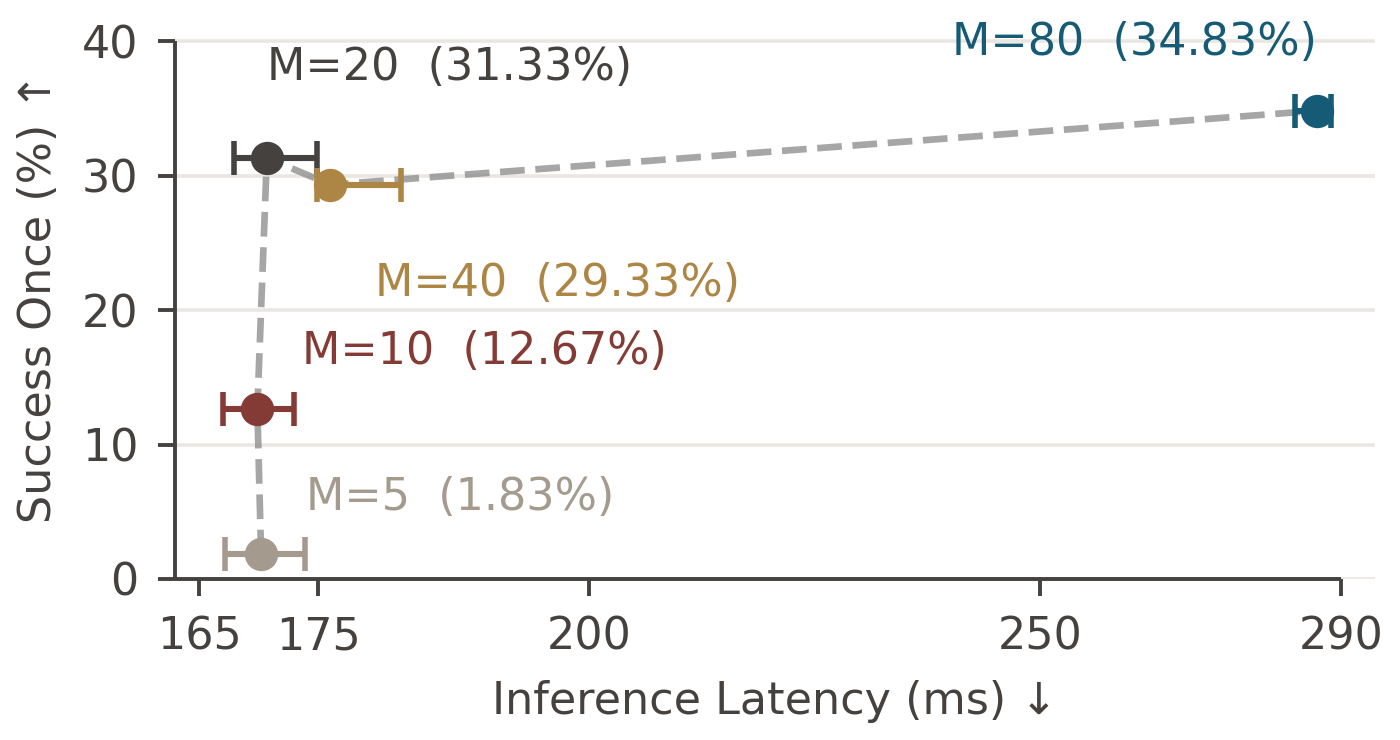}
  \caption{Success--latency trade-off across vocabulary sizes $M$. Success is averaged over Pick Apple and Pick Bowl; latency is the median batch-one inference time, with error bars indicating the 5th--95th percentiles.}
  \label{fig:k_sweep}
  \vspace{-3ex}
\end{figure}

\noindent \textbf{Impact of Action Chunk Length ($H$).} We evaluate \textsc{MobiGuide} against a deterministic $L_1$ regression baseline across commit lengths $H \in \{1,2,5,8,10\}$ in Fig.~\ref{fig:chunk_impact}. Increasing $H$ reduces the frequency of action-chunk prediction but extends the interval between replanning updates. \textsc{MobiGuide} achieves higher success rates across all evaluated commit lengths and exhibits a smaller overall performance decline. Specifically, from $H=1$ to $H=10$, its success rate decreases from $53.0\%$ to $36.0\%$ (a decline of $17.0$ percentage points), compared with $42.7\%$ to $8.7\%$ for the $L_1$ baseline (a decline of $34.0$ percentage points). These results indicate that \textsc{MobiGuide} better tolerates less frequent replanning on this task, retaining higher success when more actions are committed between policy queries.

\noindent\textbf{Impact of Vocabulary Size.} Fig.~\ref{fig:k_sweep} evaluates the success--latency trade-off across vocabulary sizes $M$, averaging success over Pick Apple and Pick Bowl. Increasing $M$ from 5 to 20 raises success from 1.83\% to 31.33\%, while median batch-one inference latency remains near $170\,\mathrm{ms}$, yielding substantial gains at little latency cost. Increasing the vocabulary to $M=40$ does not improve success (29.33\%), whereas $M=80$ achieves the highest success of 34.83\% at $286.5\,\mathrm{ms}$, with more candidates to refine and score. We therefore use $M=20$ as a practical balance: $M=80$ provides an additional 3.50 percentage points of success at approximately 68\% higher latency. These results highlight the importance of evaluating vocabulary size jointly with inference cost.

\noindent\textbf{Action Attenuation.} We evaluate residual correction under 10\% and 20\% attenuation of arm, torso, and base commands at $H=5$. As shown in Table~\ref{tab:attenuation}, residual correction improves success in all 12 attenuated settings. These results suggest that residual correction consistently benefits execution under reduced actuation across the evaluated tasks.

\subsection{Visualization}
Fig.~\ref{fig:demo_vis} visualizes \textsc{MobiGuide}'s trajectory predictions during continuous execution on \textsc{Pick Bowl} and \textsc{Place Apple}. At each timestep, predicted future end-effector poses are projected into the robot's visual workspace. White points show candidate trajectories reconstructed from anchored proposals, while colored points indicate the trajectory selected by the scoring head. Candidates span multiple approach directions early in execution and concentrate around fine-grained manipulation near the target. This illustrates how \textsc{MobiGuide} maintains multiple trajectory hypotheses while selecting one for control.

\section{CONCLUSIONS}

\label{sec:conc}
We introduce \textsc{MobiGuide}, an efficient generative framework for dynamic mobile manipulation that captures multimodal expert behavior while meeting real-time whole-body control constraints. Its anchor-guided diffusion refines demonstration-derived anchors with truncated denoising, achieving latency competitive with deterministic policies. A lightweight scoring head selects context-compatible anchors, while residual correction mitigates long-horizon open-loop drift. Robustness remains bounded by the coverage and diversity of offline demonstrations. Future work could incorporate reinforcement learning to improve test-time self-correction and out-of-distribution recovery.


\bibliographystyle{IEEEtran}

\bibliography{references_compact}

\end{document}